\documentclass[10pt,conference,compsocconf]{IEEEtran}

\usepackage{cite}
\usepackage{amsmath,amssymb,amsfonts}
\usepackage{textcomp}
\usepackage{xcolor}
\usepackage{booktabs} 
\usepackage{algorithm}
\usepackage{subcaption}
\usepackage{float}
\usepackage{graphicx}
\usepackage{paralist}
\usepackage{grffile}
\usepackage{hyphenat}
\usepackage{cleveref}
\usepackage{tabularx}

\usepackage{setspace}
\usepackage[noend]{algorithmic}

\numberwithin{subcase}{case}

\definecolor{editcolor}{RGB}{255.0, 127.0, 0.0}

\newenvironment{Algorithm}[1][tbh]%
  {\centering
  \begin{minipage}{#1}
  \begin{algorithm}[H]}%
  {\end{algorithm}
  \end{minipage}\par
  \vspace{\belowdisplayskip}}

\def\BibTeX{{\rm B\kern-.05em{\sc i\kern-.025em b}\kern-.08em
    T\kern-.1667em\lower.7ex\hbox{E}\kern-.125emX}}

\begin{document}

\title{Entropy-Isomap: Manifold Learning for High-dimensional Dynamic Processes}

\author{\IEEEauthorblockN{Frank Schoeneman}
\IEEEauthorblockA{\textit{Computer Science \& Engineering} \\
\textit{University at Buffalo}\\
Buffalo, New York, USA\\
fvschoen@buffalo.edu}

\and
\IEEEauthorblockN{Varun Chandola}
\IEEEauthorblockA{\textit{Computer Science \& Engineering} \\
\textit{Computational and Data-enabled Science \& Engineering} \\
\textit{University at Buffalo}\\
Buffalo, New York, USA\\
chandola@buffalo.edu}
\and
\IEEEauthorblockN{Nils Napp}
\IEEEauthorblockA{\textit{Computer Science \& Engineering} \\
\textit{University at Buffalo}\\
Buffalo, New York, USA \\
nnapp@buffalo.edu}
\and
\IEEEauthorblockN{Olga Wodo}
\IEEEauthorblockA{\textit{Materials Design \& Innovation} \\
\textit{University at Buffalo}\\
Buffalo, New York, USA \\
olgawodo@buffalo.edu}
\and
\IEEEauthorblockN{Jaroslaw Zola}
\IEEEauthorblockA{\textit{Computer Science \& Engineering} \\
\textit{Biomedical Informatics} \\
\textit{University at Buffalo}\\
Buffalo, New York, USA \\
jzola@buffalo.edu} }

\maketitle

\begin{abstract}
Scientific and engineering processes deliver massive high-dimensional data sets that are generated as non-linear transformations of an initial state and few process parameters. Mapping such data to a low-dimensional manifold facilitates better understanding of the underlying processes, and enables their optimization. In this paper, we first show that off-the-shelf non-linear spectral dimensionality reduction methods, e.g., Isomap, fail for such data, primarily due to the presence of strong temporal correlations. Then, we propose a novel method, {\sf Entropy-Isomap}, to address the issue. The proposed method is successfully applied to large data describing a fabrication process of organic materials. The resulting low-dimensional representation correctly captures process control variables, allows for low-dimensional visualization of the material morphology evolution, and provides key insights to improve~the~process.
\end{abstract}



\section{Introduction}

The vast majority of the current big data, especially coming from high-performance high-fidelity numerical simulations and high-resolution scientific instruments, is a result of complex non-linear processes. While these non-linear processes can be characterized by low-dimensional sub-manifolds, the actual observable data they generate is high-dimensional. This fact means that the resulting data can be represented more concisely by using a latent state, and more importantly, that physical processes described by the observed data might be better understood by discovering their underlying low dimensionality. 

\begin{figure*}[t]
\centering
\scriptsize
\begin{tabular}[t]{lcccc}
\raisebox{\height}{$\Gamma_1 = \Gamma(\phi = 0.6, \chi = 2.2)$} & \includegraphics[scale=0.7]{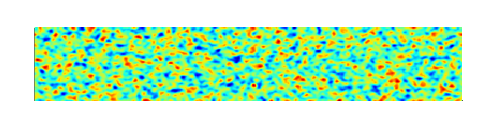} & \includegraphics[scale=0.7]{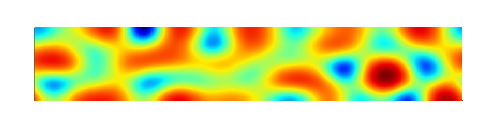} & \includegraphics[scale=0.7]{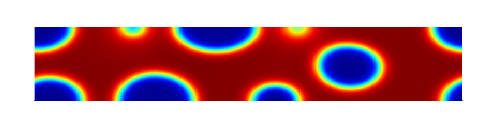} & \includegraphics[scale=0.7]{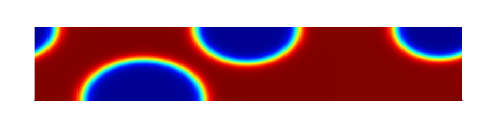}\\

\raisebox{\height}{$\Gamma_2 = \Gamma(\phi = 0.6, \chi = 3.0)$} & \includegraphics[scale=0.7]{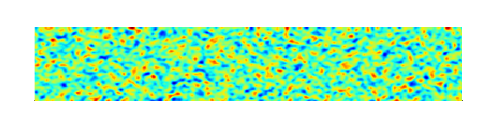} & \includegraphics[scale=0.7]{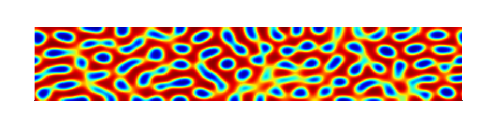} & \includegraphics[scale=0.7]{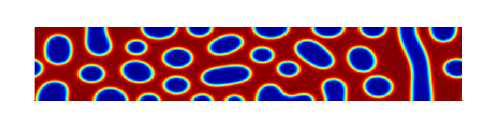} & \includegraphics[scale=0.7]{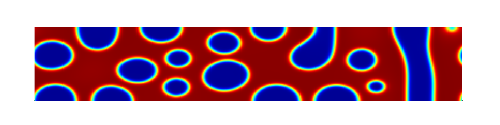}\\

\raisebox{\height}{$\Gamma_3 = \Gamma(\phi = 0.5, \chi = 3.0)$} & \includegraphics[scale=0.7]{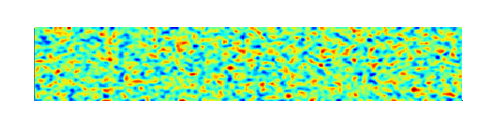} & \includegraphics[scale=0.7]{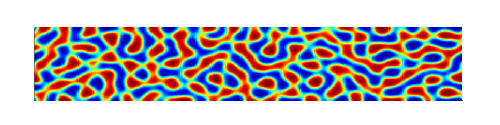} & \includegraphics[scale=0.7]{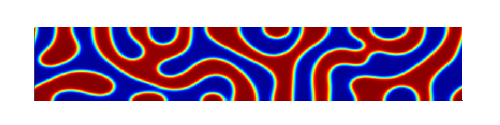} & \includegraphics[scale=0.7]{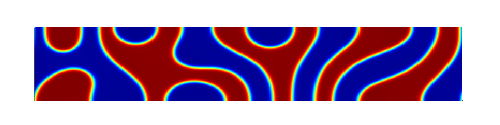}\\

  & \textsf{Time step} $0$ & \textsf{Time step} $20$  & \textsf{Time step} $80$ & \textsf{Time step} $150$\\

\end{tabular}
\caption{Sample of high-dimensional dynamic process data with three trajectories and 12 points. Each trajectory $\Gamma_I$ corresponds to different variant of organic thin film fabrication process (described by parameters $\phi$ and $\chi$). Each image is a high-dimensional point capturing material morphology (different colors represent different types of polymer making the material). Each trajectory describes morphology evolution over time. (Please view in color).}\label{fig:morphs}
\end{figure*}

Our focus in this work is on the second point, specifically, we propose a novel method for dimensionality reduction of {\it process data}. Here process data means any data that represents evolution of some process states over time (see for example Fig.~\ref{fig:morphs}). While such data are ubiquitous, they are challenging for current dimensionality reduction techniques. This is because the input data sets are large, as each sample of a process delivers a time series of high dimensional points, the underlying processes are highly non-linear, which rules out many methods that otherwise would be computationally feasible, and finally, the individual data points are sampled in a highly correlated way, which can easily confuse many dimensionality reduction~techniques.

In our prior work, we have developed {\sf S-Isomap}, a spectral dimensionality reduction technique for non-linear big data streams~\cite{Schoeneman2017} that addresses two of the above challenges. The method can efficiently and reliably handle large non-linear data sets, but assumes that the input data is weakly correlated. Consequently, it fails when applied directly to process data. {\sf S-Isomap} has been derived from the standard Isomap algorithm~\cite{Tenenbaum2000}, which is frequently used and favored in the scientific computing data analysis~\cite{Lim2003,Dawson2005,Zhang2006,Rohde2008,Ruan2014,Strange2014,Samudrala2015}. Unfortunately, while there is some prior work on applying Isomap to spatio-temporal data~\cite{Jenkins2004}, the focus has been on segmentation of data trajectories rather than discovering a continuous latent state. To the best of our knowledge, currently there are no spectral methods that can handle high-dimensional process data.

The current work is motivated by the need to analyze massive and high-dimensional data sets generated from highly non-linear differential equations modeling material morphology evolution during fabrication process of organic thin films (see Section~\ref{sec:datagen}). The fabrication of organic thin films is a key factor controlling properties of organic electronics, including transistors, batteries, and displays, but is computationally expensive and difficult to model precisely. Depending on the fabrication parameters, different process trajectories are possible, leading to different material properties. Scientists and engineers are interested in using dimensionality reduction on the resulting big data to explore the material design space, and optimize the fabrication to make devices with desired properties.

In this paper, we first show that linear techniques, such as Principal Component Analysis (PCA)~\cite{Pearson1901}, 
overestimate the latent dimension of the process, and that dimensionality reduction techniques that assume uniformity in sampling, including non-linear strategies, fail due to the highly correlated nature of process data. We hypothesize that the poor performance of non-linear techniques is related to a lack of {\em mixing} (or {\em cross-talk}) between different trajectories of a process, and present remedy based on data resampling. Our main contribution is the concept of {\em neighborhood entropy} of a point, which indicates mixing between process trajectories. We use neighborhood entropy to adaptively size the neighborhoods when computing the geodesic distance approximation in Isomap. The resulting dimensionality reduction method is both easy to implement and effective, and could likely be extended to other spectral dimensionality reduction~approaches. 

\section{Preliminaries}
\subsection{Spectral Dimensionality Reduction}\label{sec:sdr}
Spectral Dimensionality Reduction (SDR) refers to a class of methods used to identify low-dimensional structure in high-dimensional data. For a given set of points, ${\bf X}$, in a high-dimensional space~$\mathbb{R}^D$, SDR methods work by computing either top or bottom eigenvectors (eigendecomposition) of a feature matrix derived from~${\bf X}$. Here, the feature matrix, ${\bf F}$, captures structure of the data through some selected property (e.g., pairwise distances). The SDR methods rely on the assumption that there exists a function $f:\mathbb{R}^d \rightarrow \mathbb{R}^D$, $d \leq D$, that maps low-dimensional representation, $y_i \in \mathbb{R}^d$, of each data sample to the observed $x_i \in \mathbb{R}^D$. The goal then becomes to learn the inverse mapping, $f^{-1}$, that can be used to map high-dimensional $x_i$ to low-dimensional $y_i$.

The SDR methods are frequently categorized based on the assumption they make about $f$ (i.e., linear vs. non-linear). Linear methods assume that the data lie on a low-dimensional subspace $\mathbb{V}^{d}$ of $\mathbb{R}^D$, and construct a set of basis vectors representing the mapping. However, in many cases the data is complex, and the linearity assumption is too restricting. In such cases, non-linear methods work off the assumption that the data is sampled from some low-dimensional submanifold $M^{d}$, embedded within $\mathbb{R}^{D}$.

When working with the linearity assumption the most commonly used methods are PCA and Multidimensional Scaling (MDS)~\cite{Cox2000}. PCA learns the subspace that best preserves covariance. The basis vectors learned by PCA, known as principal components, are the directions along which the data has highest variance. The data $\textbf{X}$ can be transformed by first computing the covariance matrix, $\textbf{F}_{D\times D}$. In the case of MDS, the feature matrix $\textbf{F}_{n\times n}$ contains some pairwise relationships between data points in $\textbf{X}$. When these relationships are Euclidean distances, the result is equivalent to that of PCA, and this is known as classical MDS. Spectral decomposition of \textbf{F} in both methods yields eigenvectors \textbf{Q}. Taking the top $d$ eigenvectors, the data can be mapped to low-dimensions as $\textbf{Y}$ by the transformation \textbf{Y} = \textbf{X}$\textbf{Q}_{d}$.

\begin{Algorithm}[0.45\textwidth]
    \caption{\textsc{Isomap}}
	\begin{algorithmic}[1]
    \REQUIRE $\textbf{X}$, $k$
    \ENSURE $\textbf{Y}$
    \STATE $\textbf{D}_{n\times n}$ $\leftarrow$ \textsc{PairwiseDistances}($\textbf{X}$) 
    \STATE $\textbf{G}_{n\times n}$ $\leftarrow \infty$
    \FOR{$x_i \in \textbf{X}$} 
        \STATE {\bf kNN} $\leftarrow$ \textsc{KNN}($x_i$, $\textbf{X}$, $k$)
        \FOR{$x_j \in $ {\bf kNN}}
            \STATE $\textbf{G}_{i, j} \leftarrow \textbf{D}_{i, j}$
        \ENDFOR
    \ENDFOR

    \STATE $\textbf{F}_{n\times n}\leftarrow $ \textsc{AllPairsShortestPaths}($\textbf{G}$) 
    \STATE $\textbf{Y} \leftarrow \textsc{MDS}(\textbf{F})$
    \RETURN{$\textbf{Y}$}
  \end{algorithmic}
  \label{alg:isomap}
\end{Algorithm}

In cases when data is assumed to be generated by some non-linear process, both PCA and MDS are not robust enough to learn the inverse mapping $f^{-1}$. Although variants of PCA have been proposed to address such situations (e.g. Kernel PCA~\cite{Scholkopf1998}), the most common approach is to use Isomap~\cite{Tenenbaum2000}. Isomap constructs feature matrix by approximating distances between input points along the manifold $M^{d}$, and then proceeds as regular MDS. This is accomplished in four steps, as shown in Algorithm~\ref{alg:isomap}. First all $n^{2}$ pairwise distances are computed for points in $\textbf{X}$. Then geodesic distances along manifold are approximated (lines 3--6) by first constructing a neighborhood graph, \emph{G}, where each point $x \in \textbf{X}$ is adjacent to its $k$-nearest-neighbors, and then by computing shortest paths between all points in~\emph{G} (line 7). The resulting geodesics approximations are contained in the feature matrix \textbf{F}, which is processed by MDS to yield the final low-dimensional transformation.

\subsection{Dynamic Process Data}\label{sec:dynamicprocess}

When dataset $\textbf{X} \in \mathbb{R}^D$ represents a dynamic process, points in $\textbf{X}$ are partitioned into $T$ trajectories, $\Gamma_1$, $\Gamma_2$,..., $\Gamma_T$. Each trajectory $\Gamma_I$ is given by a $\tau$-parameterized sequence of $m_I$ data points. In other words, $\Gamma_I$ = ($x_I(\tau_1)$, $x_I(\tau_2)$,\ldots, $x_I(\tau_{m_I})$), where $\tau_i < \tau_j$ when $i<j$. Parameter $\tau$ usually denotes time, and trajectory $\Gamma_I$ can be a function of one or more additional~parameters. 

In this work, we investigate the use of SDR methods in the analysis of dynamic processes. As a representative example, we use numerical simulation of material morphology evolution during fabrication of organic thin film~\cite{WodoBaskar2012-CmpMatSc}. The input data consists of trajectories $\Gamma = \Gamma(\phi, \chi)$, where each trajectory is a function of two variables corresponding to two fabrication parameters (see Fig.~\ref{fig:morphs}): $\phi$,~which denotes blend ratio of polymers making organic film, and~$\chi$, which denotes strength of interaction between these polymers. Each data point $x(\tau)$ is an image representing one morphology snapshot generated by complex non-linear differential equation solver modeling morphology evolution in time. Each image is then represented by a high-dimensional vector in $\mathbb{R}^D$, obtained by simple processing of image pixels. Example morphologies from selected trajectories are shown in Fig.~\ref{fig:morphs}, and we give detailed description of the data and the data generation process in Section~\ref{sec:app}.

The main challenge in analyzing the temporal morphology evolution data comes from the inherent bias in the exploration of possible states of the fabrication process. In the essence, sampling in $\tau$ (i.e., time) is commonly unbalanced meaning much more dense than in parameters $\phi$ or $\chi$. This is because the computational cost of executing solver to generate a single trajectory (i.e., sampling in $\tau$) is prohibitive to allow for the exhaustive sampling in the space formed by parameters $\phi$ and $\chi$. Furthermore, data points in the same trajectory have high temporal correlation, which is reflective of how morphologies evolve. These factors strongly influence connectivity of the neighborhood graph, $G$, and in turn affect approximation of the manifold distances.

\section{Challenges in Using SDR with Dynamic Process Data}

The standard off-the-shelf approach to perform dimensionality reduction on large data is PCA. However, if the method is applied without taking into consideration the underlying assumption of data linearity, it delivers highly misleading results. Here we study effectiveness of both PCA and Isomap when dealing with dynamic process data.

\subsection{Effectiveness of State of the Art SDR Methods}

A reliable way of determining the quality of the low dimensional representation (mapping) produced by each method is to compare the original data $\textbf{X}$ in $\mathbb{R}^D$ with the mapped data $\textbf{Y}$ in $\mathbb{R}^d$, by computing the {\em residual variance}. The process of computing residual variance for PCA differs from Isomap, but the values are directly comparable. 

In PCA, each principal component (PC) explains a fraction of the total variance in the dataset. If we consider $\lambda_{i}$ as the eigenvalue corresponding to the $i^{th}$ PC and $\vert\Lambda\vert$ as the total energy in the spectrum, i.e., $\vert \Lambda \vert = \sum_{i=1}^D \lambda_i$, then the variance explained by the $i^{th}$ PC can be computed as $\frac{\lambda_{i}}{\vert\Lambda\vert}$. The residual variance can be calculated~as: 
\begin{equation}\label{resvardef}
R = 1 - \sum_{i=1}^{d}\frac{\lambda_{i}}{\vert\Lambda\vert}.
\end{equation}

In the Isomap setting, residual variance is computed by comparing the approximate pairwise geodesic distances, computed in \emph{G} represented by matrix $\textbf{D}_{G}$ (recall that $G$ is a neighborhood graph), to the pairwise distances of the mapped data $\textbf{Y}$, represented by matrix $\textbf{D}_{Y}$:
\begin{equation}\label{Iso_resvardef}
R = 1 - \rho(\textbf{D}_{G}, \textbf{D}_{Y})^{2}.
\end{equation}
Here, $\rho$ is the standard linear correlation coefficient, taken over all entries of ${\bf D}_G$ and ${\bf D}_Y$.

\begin{figure}[ht]
   \centering
   \includegraphics[]{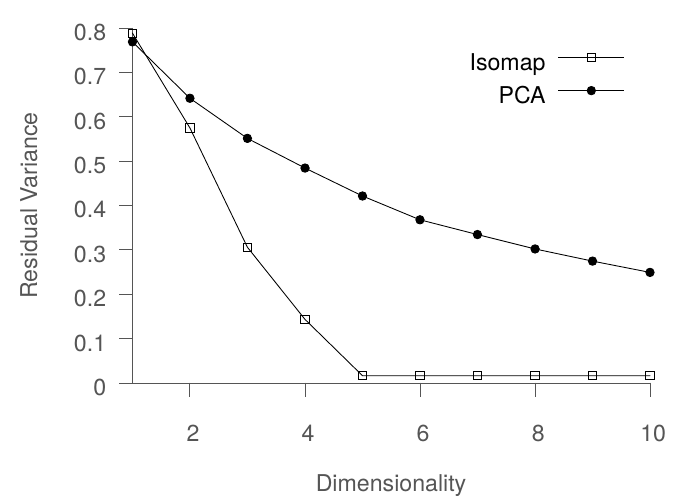}
   \caption{Isomap and PCA run on data with six trajectories for $\chi = 3.0$ and $\phi\in\{0.50, 0.52, 0.54, 0.56, 0.58, 0.6\}$. The quality of the Isomap manifold and PCA subspace are assessed using residual~variance.}
   \label{chi300scree}
\end{figure}

In the first step of our analysis, we compared the residual variance obtained using PCA and Isomap on the material morphology evolution process data (see Sections~\ref{sec:dynamicprocess} and~\ref{sec:app}) consisting of six different trajectories, each trajectory corresponding to a unique configuration of pair $\phi$ and $\chi$. Figure~\ref{chi300scree} summarizes our findings for PCA and Isomap. From the figure, we can see that PCA in unable to learn an effective low-dimensional mapping. In fact, while Isomap is able to explain about 70\% of the variance using 3 dimensions, PCA requires more than 9 dimensions. Here we note that the ability to explain most of the information in the data in two or three dimensions is highly desired by domain experts as it permits data visualization and exploratory analysis.

\begin{figure*}[htbp!]
 	\begin{subfigure}[t]{0.32\textwidth}
     	\includegraphics[width=.99\textwidth,trim={3.75cm 1.25cm 1.75cm 1.75cm},clip]{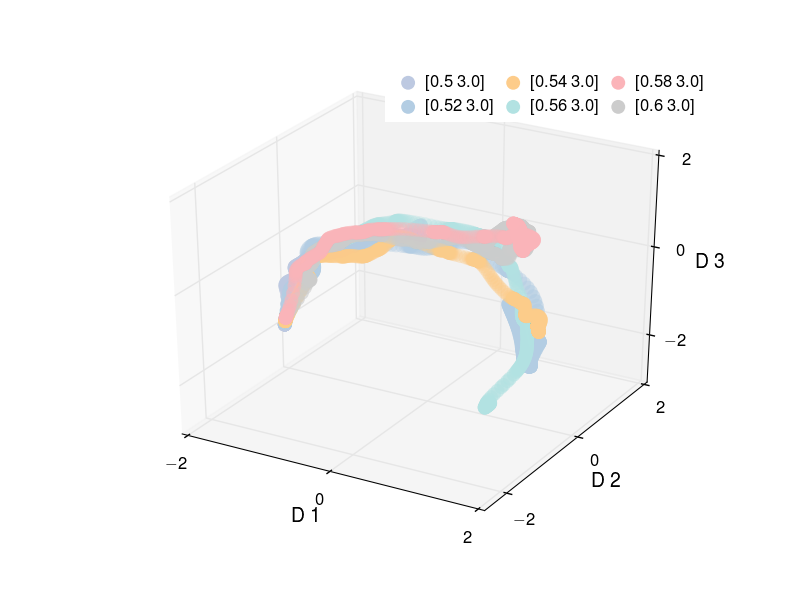}
         \caption{\label{chi300_fftR_pca}}
 	\end{subfigure}
    \begin{subfigure}[t]{0.32\textwidth}
    	\includegraphics[width=0.99\textwidth,trim={2.25cm 1.25cm 1.75cm 1.75cm},clip]{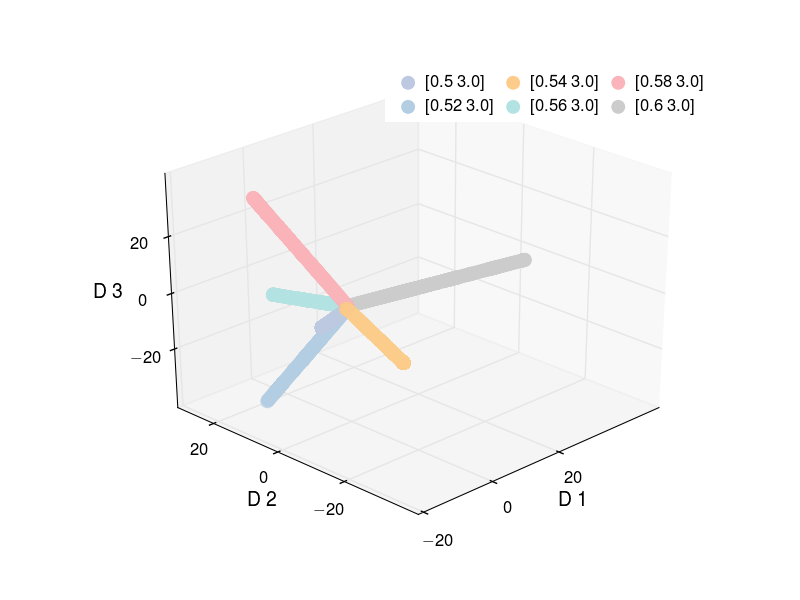}
        \caption{\label{chi300_fftR_iso}}
    \end{subfigure}
    \begin{subfigure}[t]{0.32\textwidth}       
	\includegraphics[width=0.99\textwidth,trim={3.70cm 1.5cm 1.75cm 1.5cm},clip]{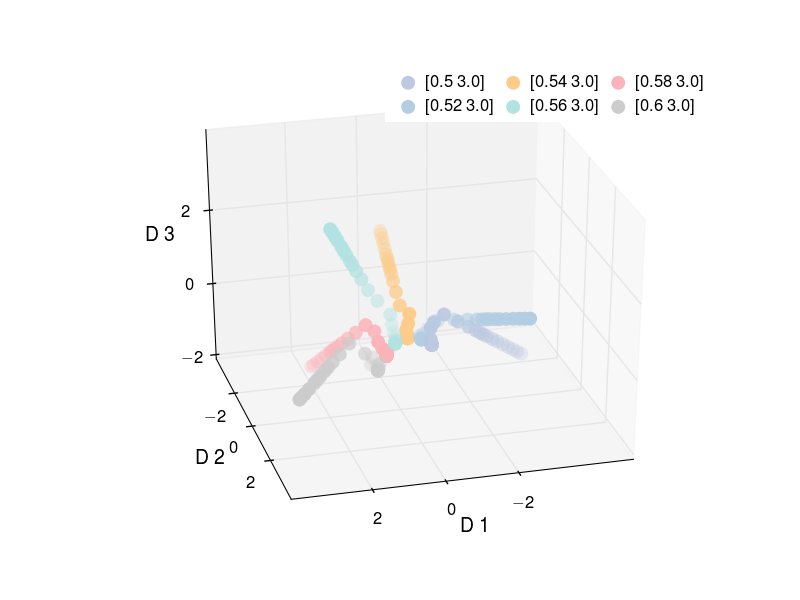} 
    \caption{\label{chi300_fftR_early_iso3d}}
    \end{subfigure}
    \caption{
    Six trajectories with fixed $\chi = 3.0$ and variable $\phi\in\{0.50, 0.52, 0.54, 0.56, 0.58, 0.6\}$ were selected to learn mapping and transform the data to 3-dimensions using (a) PCA (b) Isomap with $k=8$ (c) Isomap with $k=8$ using only the first 30 time steps of each pathway. (Please view in color).
    }
    \label{chi300_sdr_figs}
\end{figure*}


To visualize the data, we used both PCA and Isomap to map the data to $d=3$ dimensions. The results are shown in Fig.~\ref{chi300_fftR_pca} and~\ref{chi300_fftR_iso}. For PCA, one of the dimensions ($D1$) describes the time aspect of the process evolution. However, the PCA visualization does not offer additional insights into the process, which we attribute primarily to the PCA's inability to capture non-linearities.

Since Isomap outperforms PCA in terms of residual variance, it is expected that the 3-dimensional data obtained from Isomap would offer more meaningful insights. However, as shown in Fig.~\ref{chi300_fftR_iso}, all trajectories diverge from one another in $3$-dimensions and there is no reasonable interpretation of the empty space. This indicates that the standard application of Isomap is inadequate when working with parameterized high-dimensional time series data. We note that we obtained equally unsatisfactory results with other methods, including t-SNE~\cite{Maaten2008} and LLE~\cite{Roweis2000}.

\subsection{Standard Isomap and Dynamic Process Data}

To further study the reason behind Isomap performance, we focus on the initial stage of the trajectories, where the morphologies are expected to evolve in a similar fashion. This is reflected in the Isomap visualization in Fig.~\ref{chi300_fftR_iso}, where all trajectories appear to start from a common point in the 3-dimensional space and then diverge.

\begin{figure*}[htbp!]
	\begin{subfigure}[t]{0.49\textwidth}
    \centering\captionsetup{width=.95\linewidth}%
   		\includegraphics[]{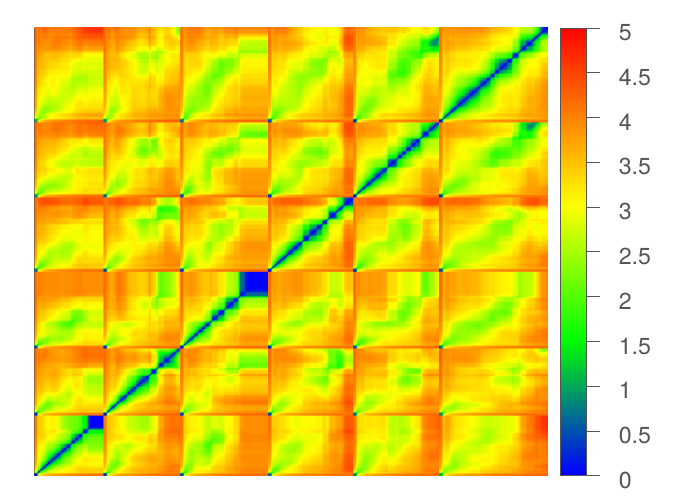}
   		\caption{\label{chi300_fftR_pwdheat}\
        Pairwise distance matrix for all data grouped by $\phi$ and ordered by time step along both axes. 
        Distances in subblocks along the main diagonal denote inter-point distances within a fixed $\phi$-value trajectory. 
        Off-diagonal subblocks highlight distance between points lying on disjoint trajectories.}
	\end{subfigure}
	\begin{subfigure}[t]{0.49\textwidth}
    	\centering\captionsetup{width=.95\linewidth}
   		\includegraphics[]{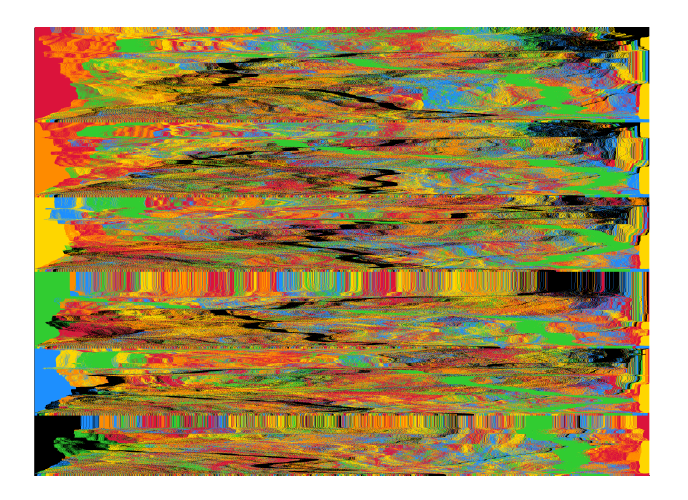}
   		\caption{\label{chi300_fftR_sortedpwdheat}\
        Distance matrix with rows grouped by $\phi$ and ordered by time step. Entries in row $i$ are sorted by increasing distance from $x_i$ and colored according with their $\phi$ value. 
        Clusters of similar color nearest the left edge reflect $k$-NN having common $\phi$-value for sufficiently large $k$.}
        \end{subfigure}
        \caption{Pairwise distances of all points points with $\chi = 3.0$ and from six trajectories for $\phi\in\{0.50, 0.52, 0.54, 0.56, 0.58, 0.6\}$ visualized in two ways. (Please view in color).}
        \label{chi300_heatmaps}
\end{figure*}

We applied Isomap on only the early stage data represented by the first $30$ time steps of each trajectory (threshold selected by the domain expert). The results are shown in Fig.~\ref{chi300_fftR_early_iso3d}, where we can clearly observe that the early data points for all trajectories cluster together before quickly diverging. This leads us to the first key observation of this paper: When dealing with dynamic process data, in which the data points exhibit a strong temporal correlation within the trajectory to which they belong, but are different from data points that belong to other trajectories, Isomap cannot capture the relationships across different trajectories. Thus, the resulting mapping is dominated by the time dimension, as can be seen in Fig.~\ref{chi300_fftR_iso}. This behavior can be attributed to how neighbors are selected for each point (see Algorithm~\ref{alg:isomap}, line 4). To better illustrate the point, consider Fig.~\ref{chi300_fftR_pwdheat}, which shows the matrix, ${\bf D}$, containing the distance between every pair of points, with rows and columns ordered by trajectory and time. In Fig.~\ref{chi300_fftR_pwdheat}, the same row ordering is retained, however, each row contains the sorted distances of the corresponding point to all points in the dataset, and colored by the trajectory to which they belong. Both figures show that for the majority of the points, the first several nearest neighbors are always from the same trajectory. This is problematic, because the ability of Isomap to learn an accurate description of the underlying manifold, depends on how well the neighborhood matrix captures the relationship {\em across} the trajectories. We refer to this relationship as {\em cross-talk}, or {\em mixing} among the trajectories. For any given point, the desired effect would be that the nearest neighborhood set contains points from multiple trajectories. However, the sorted neighborhood matrix indicates a lack of mixing, which essentially means that the Isomap algorithm does not consider information from other trajectories, when learning the shape of the manifold in the neighborhood of one~trajectory.

\subsection{Quantifying Trajectory Mixing}

To better assess the quality of neighborhoods and understand the mixing of trajectories, we use the information-theoretic notion of entropy. For a given point $x$, let $p_i$ be the fraction of $k$ closest neighbors of $x$ that lie on the trajectory $\Gamma_i$. Then, the entropy of the $k$-neighborhood of point $x$ is calculated as:
\begin{equation}\label{entropy_eqn}
H^k_x = \sum_{p_{i}\neq 0}{-p_{i} \log_2 p_{i}}
\end{equation}

Similarly, we can define the $k$-neighborhood entropy for a trajectory $\Gamma$, as the average of $k$-neighborhood entropy for all points on $\Gamma$.

When the neighborhood entropy of a point is high, its nearest neighbors are uniformly distributed across all trajectories (high level of mixing). On the other hand, if the entropy of a point is low, its nearest neighbors mostly lie on a single trajectory (low level of mixing). Thus, neighborhood entropy measures the mixing level across the trajectories, for a given neighborhood size, $k$.

\begin{figure}[!ht]
   \centering
   \includegraphics[]{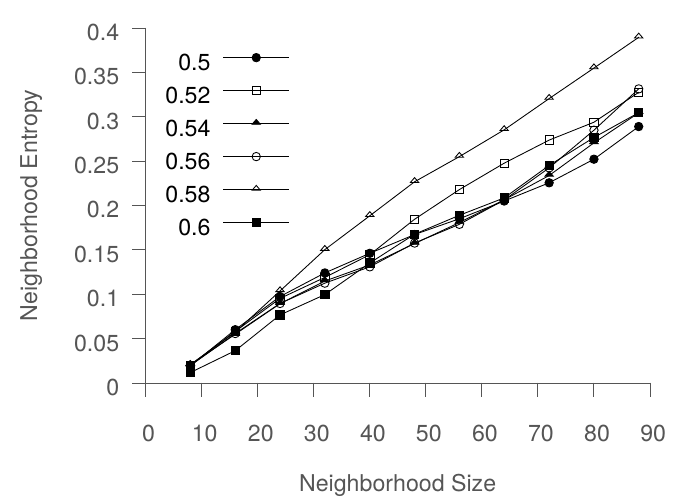}
   \caption{Neighborhood entropy of different trajectories as a function of $k$ ($\chi = 3.0$ and $\phi\in\{0.50, 0.52, 0.54, 0.56, 0.58, 0.6\}$).}
   \label{entropy_k}
\end{figure}

\subsection{Strategies for Inducing Trajectory Mixing}

One simple way to induce more trajectory mixing is to increase $k$, since this would increase the neighborhood entropy of the points. Fig.~\ref{entropy_k} shows the average neighborhood entropy for each of the six trajectories in the data set, for different values of $k$. The neighborhood entropy increases linearly with $k$, consistently for all six trajectories. Thus, for a small value of $k$, Isomap is unable to obtain a meaningful low-dimensional representation, as evident in Fig.~\ref{chi300_fftR_iso}, where $k$ was set to 8.

Figure~\ref{entropy_k} shows that using large $k$ could result in the desired level of trajectory mixing. However, as discussed in the original Isomap paper~\cite{Tenenbaum2000}, the approximation error between the true Geodesic distance on the manifold between a pair of points and the approximate distance calculated using Dijkstra's algorithm (See Algorithm~\ref{alg:isomap}) is inversely related to $k$. For large $k$, Isomap is essentially reduced to PCA, and is unable to capture the nonlinearities in the underlying manifold.

Another strategy to induce trajectory mixing is {\em subsampling}, i.e., selecting a subset of points from a given trajectory. 
However, this would result in reduction of the data, which yields poor results. Alternatively we could use {\em skipping} in the neighborhood selection, i.e., for a given point, skip the $s$ nearest points before including points in the neighborhood. 
Unfortunately, in experimenting with skipping and subsampling approaches we experienced a loss in local manifold quality or data size, respectively. Based on this and the desirability of achieving the most accurate local and global qualities, we propose an entropy-driven approach in the next section.

\section{Entropy-Isomap}
Standard Isomap does not work well for dynamic process data since data points are typically closest to other data points from the same trajectory, yet the global structure of the process depends on relations between different trajectories. When the $k$-NN neighborhoods are computed, this can result in poor mixing. Worse, how much trajectories interact can change throughout the process. For example, when trajectories come from simulations with similar initial conditions, the trajectories might interact for a while, but then diverge to explore different parts of the state space. A neighborhood size $k$ that produces good results in early stages might produce poor results later on in the process. A value of $k$  that is large enough to work for all times might include so many data points that the geodesic and Euclidean distances become essentially the same, which results in a PCA like behavior, defeating the purpose of using Isomap.

To address this situation, we propose to directly measure the amount of mixing and use it to change the neighborhood size for different data points adaptively. This mitigates the shortcomings of the two methods described in the previous section that either discard data (subsampling) or lose local information (skipping).

Figure~\ref{entropy_k} shows that neighborhood entropy increases when the next nearest neighbors are added. We propose using an entropy threshold to determine a neighborhood size $k$. This modification allows the flexibility of larger neighborhoods in regions where it is necessary or desired to force mixing between~trajectories. 

To prevent neighborhoods that are so large as to reduce Isomap to PCA, the maximum neighborhood size $M$ is left as a parameter. This check allows processing datasets which contain trajectories in poorly sampled regions of the state space without skewing the rest of the analysis, which would otherwise result in unreasonably large neighborhood sizes. 

\begin{Algorithm}[0.45\textwidth]
    \caption{\textsc{Entropy-Isomap}}
	\begin{algorithmic}[1]
    \setstretch{0.975}
    \REQUIRE $\textbf{X}$, $k$ , $\hat{H}$, $M ( = 100 )$
    \ENSURE $\textbf{Y}$ 

    \STATE $\textbf{D}_{n\times n}$ $\leftarrow$ \textsc{PairwiseDistances}($X$)
    \STATE $\textbf{G}_{n\times n}$ $\leftarrow \infty$  

    \FORALL{$x_i \in \textbf{X}$}
        \STATE $k_i \leftarrow k$
        \WHILE {$H < \hat{H}$ \textbf{and} $k_i < (M + k)$}
            \STATE $k_i \leftarrow k_i + 1$
            \STATE \textbf{kNN} $\leftarrow$ \textsc{KNN}($x_i$, $\textbf{X}$, $k$)
            \STATE $\textbf{G}_{i, j} \leftarrow \textbf{D}_{i, j}$ where $x_j \in $ \textbf{kNN}.
            \STATE $H \leftarrow$ \textsc{NeighborhoodEntropy}($x$, $k_i$, $\bar{G_i}$)
            
        \ENDWHILE
    \ENDFOR
    \STATE $\textbf{F}_{n\times n} \leftarrow$ \textsc{AllPairsShortestPaths}($\textbf{G}$)
    \STATE $\textbf{Y} \leftarrow \textsc{MDS}(\textbf{F})$
    \RETURN{$\textbf{Y}$}
  \end{algorithmic}
  \label{alg:ent_isomap} 
\end{Algorithm}

The proposed {\sf Entropy-Isomap} algorithm is shown in Figure~\ref{alg:ent_isomap}. Compared to the standard approach, the algorithm takes additional argument, the target entropy level, $\hat{H}$. This parameter is used to decide when adaptively computed neighborhoods are producing good mixing. The initial step, computing all pairwise distances for data points in ${\bf X}$, remains the same as in the standard algorithm. Then, the entropy-based neighborhood selection is performed~\mbox{(lines 3-9)}. For each point $x_i$, the algorithm proceeds with neighborhood size~$k_i$, initially equal to some default value $k$. $k_i$-nearest-neighbors are identified, and their neighborhood entropy is computed (lines 7-9). If the entropy threshold $\hat{H}$ is not satisfied, then $k_i$ is incremented (line~6), and the process repeats. Once the entropy threshold is reached, or a user-defined maximum of $M$ iterations have been performed, the process terminates. The entire process is repeated for each $x_i$, and after all neighborhoods have been identified, the algorithm continues the same way as standard Isomap (lines 10-12). We note that our presentation of the algorithm is simplified for clarity. In the practical implementation, the size of the neighborhood $k_i$ can be found via simple binary search, which further can be coupled with efficient incremental $k$-NN solver, without the need to instantiate a complete distance matrix ${\bf D}$.

\begin{figure*}[htbp!]
	\begin{subfigure}[t]{0.495\textwidth} 
    \includegraphics[width=.9\textwidth,trim={4cm 1.5cm 1.75cm 1.5cm},clip]{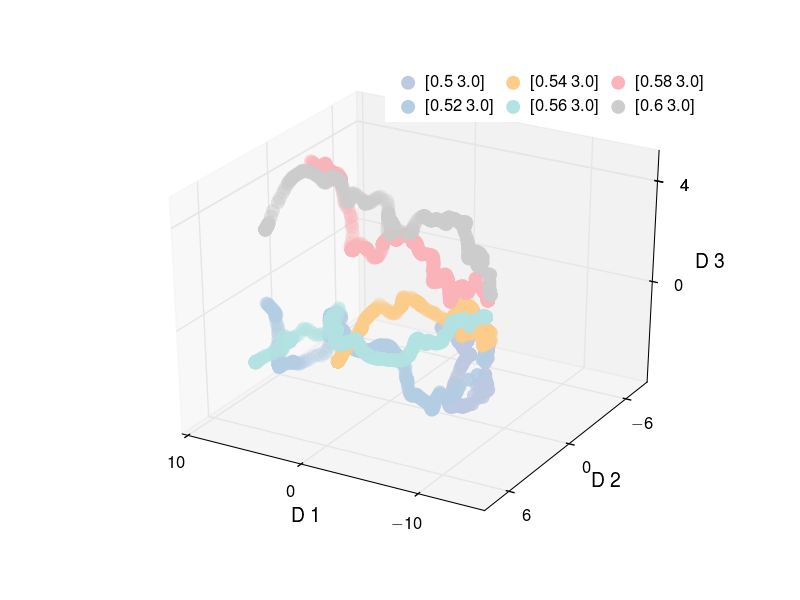} \vspace{-0.25cm}  \caption{}
\end{subfigure}
\begin{subfigure}[t]{0.495\textwidth}
    \includegraphics[width=0.99\textwidth]{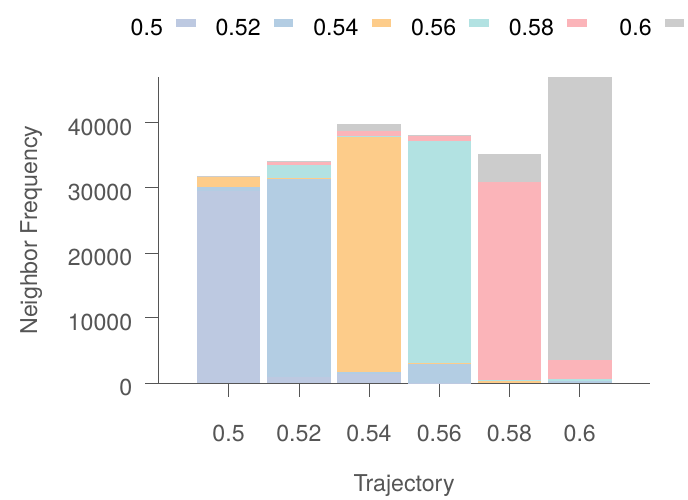}
    \vspace{-0.25cm} \caption{}
\end{subfigure} \vspace{-0.25cm}
\caption{Entropy-Isomap with $k=8$, $\hat{H}=0.3$ is run on $\chi = 3.0$ and variable $\phi\in\{0.50, 0.52, 0.54, 0.56, 0.58, 0.6\}$. (a)~The learned mapping is used to transform the data to 3-dimensions. (b)~Neighborhood cross-mixing: for each trajectory~$\Gamma$, neighbors of each point belonging to individual trajectories are aggregated and shown in stacked bar graph form. (Please~view~in~color).}
\label{eiso_iso3d_ngbr}
\end{figure*}

We applied {\sf Entropy-Isomap} to our data with $k=8$ and the maximum number of steps $M=100$. We selected this large $M$ to compute the fraction of large neighborhoods that would be required to strictly enforce mixing, in this case nearly 5\% of our data. We also varied the entropy threshold $\hat{H}$ from $0.1$ to $0.9$, to explore the effect it has on the neighborhood size distribution. Example low dimensional representation obtained by {\sf Entropy-Isomap} is presented~in~Figure~\ref{eiso_iso3d_ngbr}~(a).

\begin{figure}
    \centering 
    \includegraphics[]{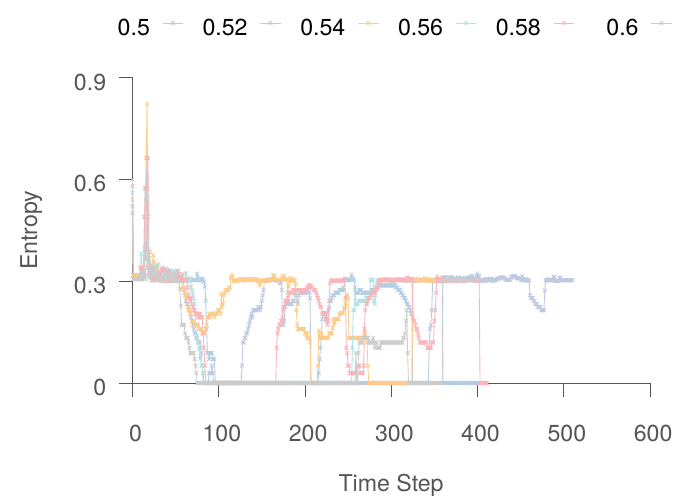}
    \caption{Entropy-Isomap with default $k=8$ and $\hat{H}=0.3$ is run on $\chi = 3.0$ and variable $\phi\in\{0.50, 0.52, 0.54, 0.56, 0.58, 0.6\}$. The entropy of the discovered neighborhood is shown for the data point at each time step. (Please view in color).}    
    \label{entropy_vs_t_eiso}
\end{figure}

We start our analysis by observing, that in the experiments high-entropy thresholds were often not reachable (see Figure~\ref{entropy_vs_t_eiso}). We believe that this is because
$k$ nearest neighbors for the majority of points are in the same trajectory (see Figure~\ref{chi300_fftR_sortedpwdheat}), which leads to skewed neighborhoods distribution. As a result, even when a satisfactory number of neighbors come from other trajectories, the entropy for the neighborhood might be low. Figure~\ref{eiso_iso3d_ngbr}~(b) shows that even when trajectories mix, the majority of neighbors are still from the same trajectory. Therefore, high entropy implies good mixing, but the converse is not necessarily true. Large neighborhoods could produce mixing, while still having low entropy. Since large neighborhoods produce poor results with Isomap, we would like to avoid them in any case. In practice, we achieved good results with entropy thresholds in the range of $\hat{H} = 0.30-0.40$, which were achievable by a large fraction of neighborhoods. This is further confirmed by experiments in Section~\ref{sec:app}.

\begin{figure}
    \centering 
    \includegraphics[]{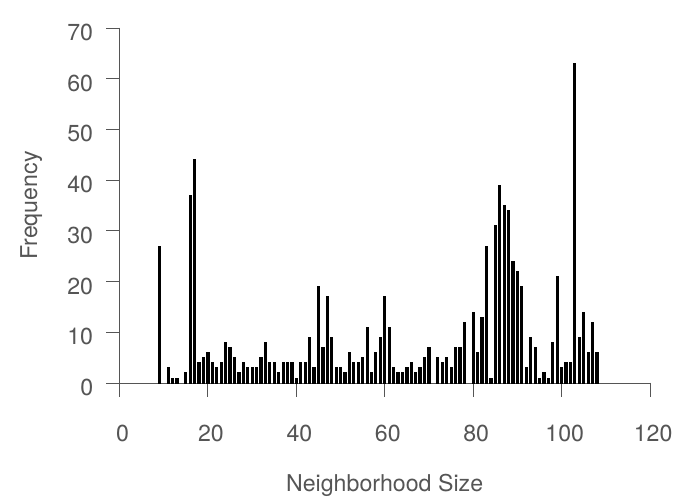}
    \caption{Entropy-Isomap with default $k=8$ and entropy threshold $\hat{H} = 0.3$ run for data with $\chi = 3.0$ and variable $\phi\in\{0.50, 0.52, 0.54, 0.56, 0.58, 0.6\}$. The distribution of selected neighborhood sizes $k_i$ that did not reach the maximum $M=100$ are shown.}
    \label{entropy_iso_kvals}
\end{figure}

When strictly enforcing entropy, the neighborhood sizes can become too large. Figure \ref{entropy_iso_kvals} shows the neighborhood size distribution for $\hat{H} = 0.30$. When such neighborhoods are included for many points in the dataset, the neighborhood graph tends toward a completely connected graph, and the  Isomap solution reduces to the PCA result. Recall that PCA is equivalent to classical MDS and that classical MDS is Isomap with $k=n-1$.

Points that produce no mixing also end up with large neighborhoods, as Entropy-Isomap tries to increase $k_i$ in order to meet the entropy threshold. These points occur when the dataset does not contain enough trajectories that pass near those particular states to produce good geodesic distance estimates. Interestingly, plotting entropy versus time in Figure~\ref{entropy_vs_t_eiso} reveals that trajectories can pass through poorly sampled parts of the state space and again ``meet up'' with other trajectories. 

The proposed methods can be used to detect trajectories that do not interact and also which regions of the state space are poorly sampled. This can be used to either remove them form the dataset or as a guide to decide where to collect more process data.

\section{Application}\label{sec:app}

The current work is motivated by the need to analyze and understand big data sets arising in the manufacturing of Organic Electronics (OE). OE is a new sustainable class of devices, spanning organic transistors~\cite{osc-transistors,osc-transistors-review}, organic solar cells~\cite{HS04,osc-book}, diode lighting~\cite{oled-review1,oled-review2}, flexible displays~\cite{displays}, integrated smart systems such as RFIDs~\cite{organic-rfdis,organic-rfdis2}, smart textiles~\cite{smartTextiles}, artificial skin~\cite{organicSkin}, and implantable medical devices and sensors~\cite{lochner2014all, zhu2014photoreconfigurable}. The critical and highly desired feature of OE is inexpensive, rapid and low-temperature roll-to-roll fabrication. However, many promising OE technologies are bottlenecked at the manufacturing stage -- more precisely, at efficiently choosing fabrication pathways that would lead to the desired material morphologies, and hence device properties.

Final properties of OE (e.g., electrical conductivity), are a function of more than a dozen material and process variables that can be tuned (e.g., evaporation rate, blend ratio of polymers, final film thickness, solubility, degree of polymerization, atmosphere, shearing stress, chemical strength and frequency of patterning substrate), leading to the combinatorial explosion of manufacturing variants. Because the standard trial-and-error approach, in which many prototypes are manufactured and tested, is too slow and cost inefficient, scientists are investigating {\em in silico} approaches. The idea is to describe the key physical processes via a set of differential equations, and then perform high-fidelity numerical simulations to capture the process dynamics in relation to input variables. Then the problem becomes to identify and simulate some initial set of manufacturing variants, and use analytics of the resulting process data to first understand the process dynamics (e.g., rate of change in domain size, or transition between different morphological classes), and then identify new promising manufacturing~variants.


\subsection{Data Generation} \label{sec:datagen}

The material morphology data analyzed in this paper, has been generated by a computational model based on the phase-field method to record the morphology evolution during thermal annealing of the organic thin films~\cite{WodoBaskar2012-CmpMatSc,WodoBaskar2011-JCP}. We focused on the exploration of two manufacturing parameters, blend ratio $\phi$ and strength of interaction $\chi$. We selected these two parameters, since they are known to strongly influence properties of the resulting morphologies. For each fabrication variant ($\phi,\chi$), we generated a series of morphologies that together formed one trajectory $\Gamma(\phi, \chi)$.

We selected the range of our design parameters $\phi=[0.5,0.6]$ and $\chi=[2.2,3.0]$ to explore several factors. First, we are interested in two stages of the process: early materials phase separation and coarsening. Moreover, we would like to explore various topological classes of morphologies. In particular, we are interested in identifying fabrication condition leading to interpenetrated structures. Finally, we seek to find the optimal annealing time that results in desired material domain sizes. In total, we generated 16 trajectories, with $180$ morpohologies on average per trajectory. Each morphology was represented as an image converted into an $40,000$-dimensional space defined by pixel composition~values.

\subsection{Results}

From the manufacturing design perspective, there are two basic aims for dimensionality reduction of morphological pathways. First, we seek to discover the common latent variables driving the dynamic process. Second, we seek to learn the geometry of manifold to device subsequent round of input parameter space exploration.  

\begin{figure*}
    \centering 
    \includegraphics[width=.24\textwidth,trim={3cm 1.5cm 1.75cm 1.5cm},clip]{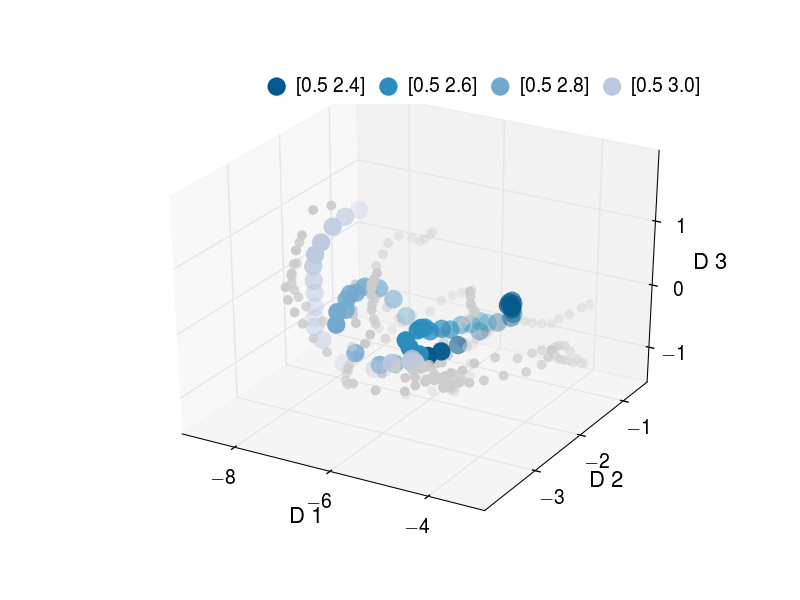}
    \includegraphics[width=.24\textwidth,trim={3cm 1.5cm 1.75cm 1.5cm},clip]{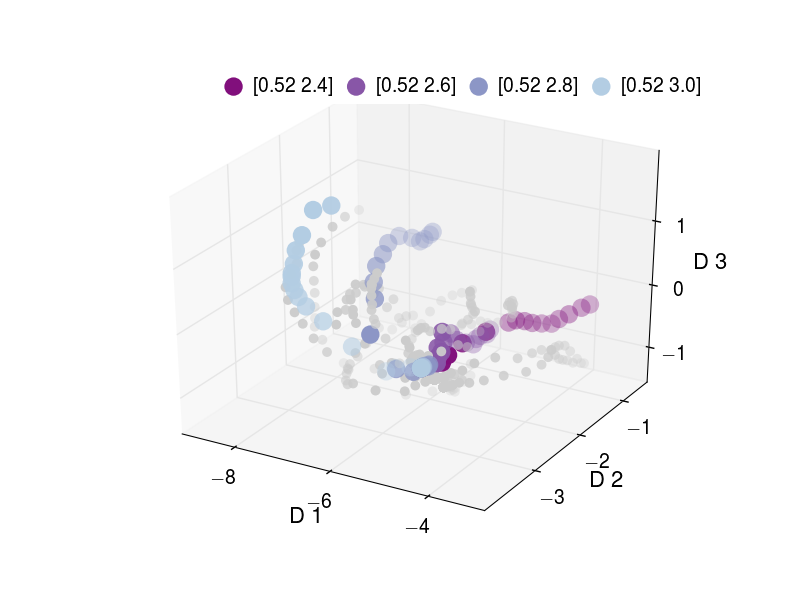}
    \includegraphics[width=.24\textwidth,trim={3cm 1.5cm 1.75cm 1.5cm},clip]{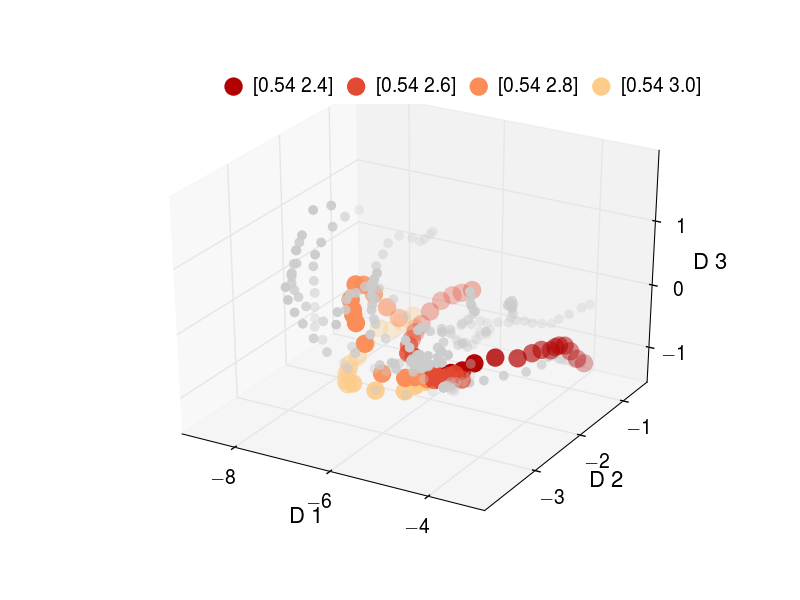}
    \includegraphics[width=.24\textwidth,trim={3cm 1.5cm 1.75cm 1.5cm},clip]{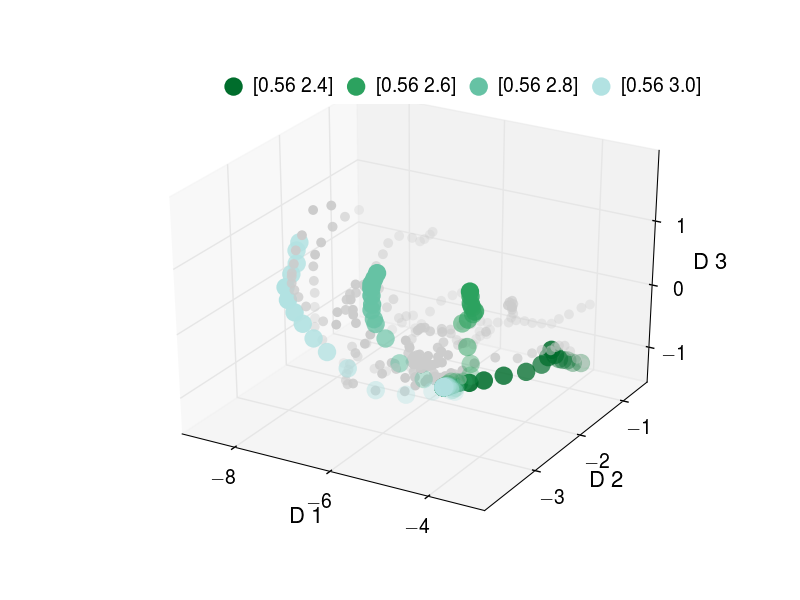}

\includegraphics[width=.24\textwidth,trim={3cm 1.5cm 1.75cm 1.5cm},clip]{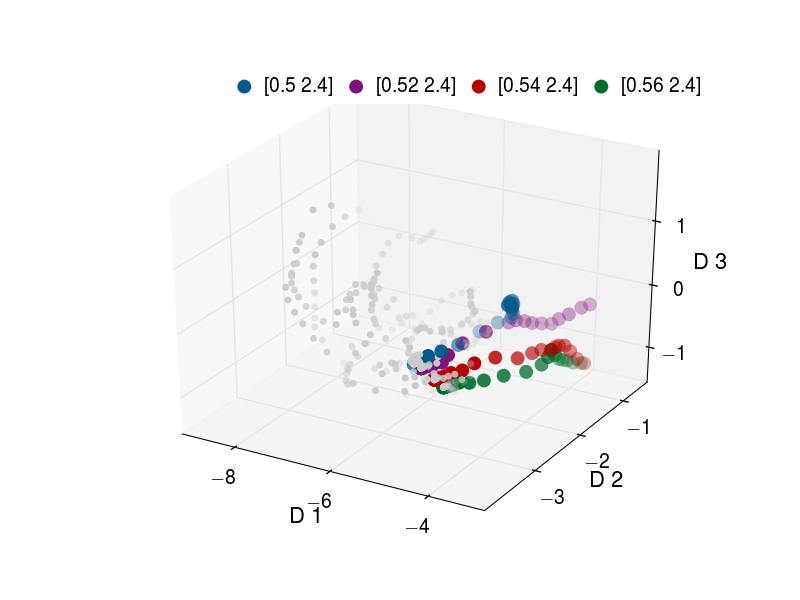}
    \includegraphics[width=.24\textwidth,trim={3cm 1.5cm 1.75cm 1.5cm},clip]{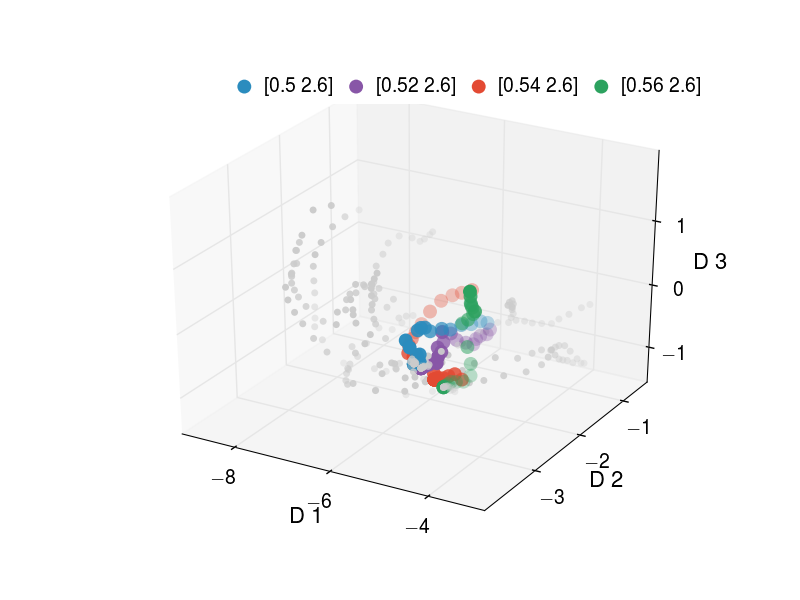}
    \includegraphics[width=.24\textwidth,trim={3cm 1.5cm 1.75cm 1.5cm},clip]{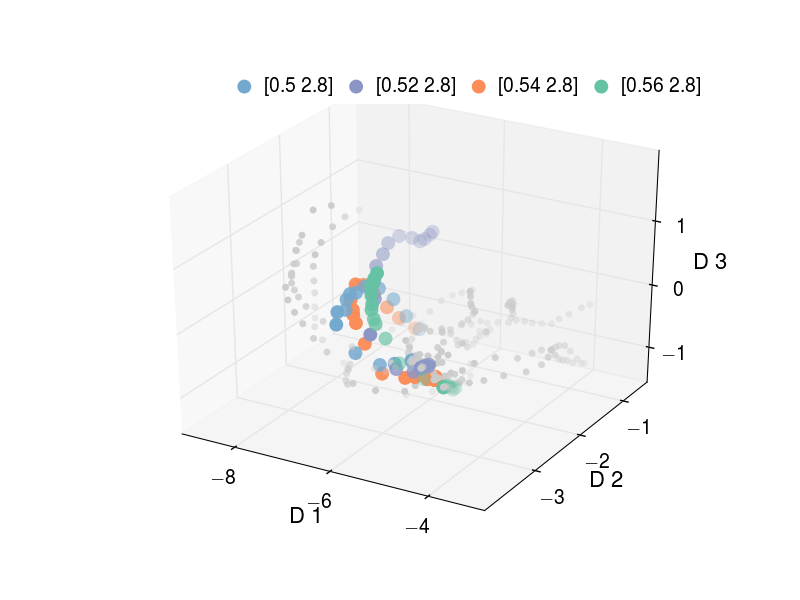}
    \includegraphics[width=.24\textwidth,trim={3cm 1.5cm 1.75cm 1.5cm},clip]{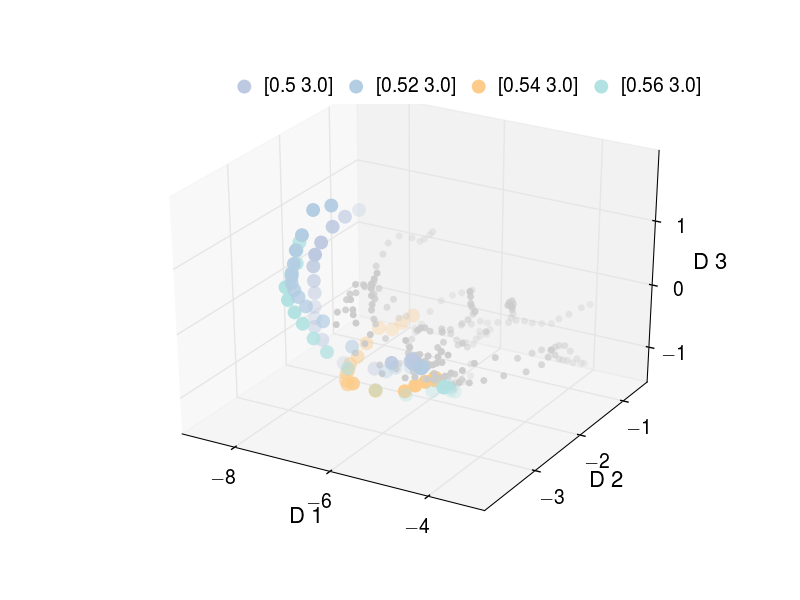}
    \caption{The manifold of early stage of the morphology evolution with first 30 points per each trajectory. To better illustrate discovered ordering by two variables we color coded the same manifold with increasing $\phi$ (top) and $\chi$ (bottom). (Please view in color).}    
    \label{results:early}
\end{figure*}

\begin{figure*}
    \centering 
    \includegraphics[width=.24\textwidth,trim={3cm 1.5cm 1.75cm 1.5cm},clip]{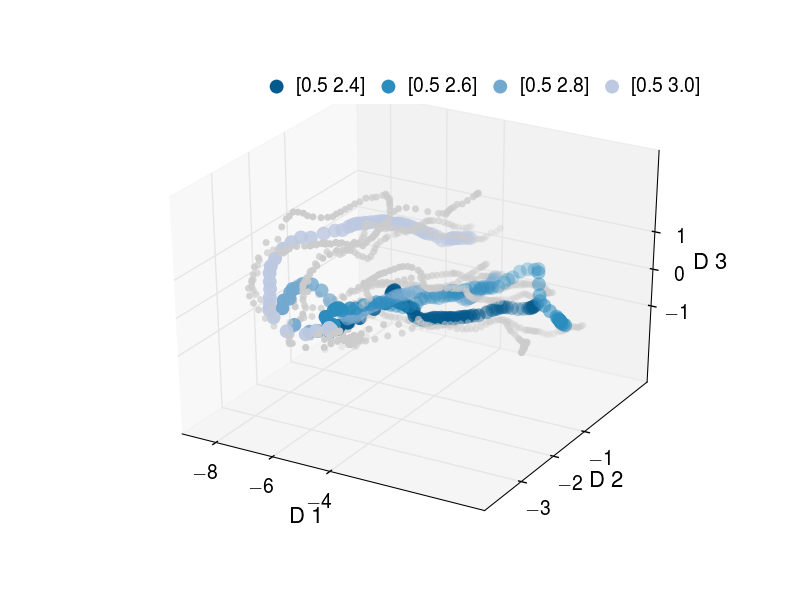}
    \includegraphics[width=.24\textwidth,trim={3cm 1.5cm 1.75cm 1.5cm},clip]{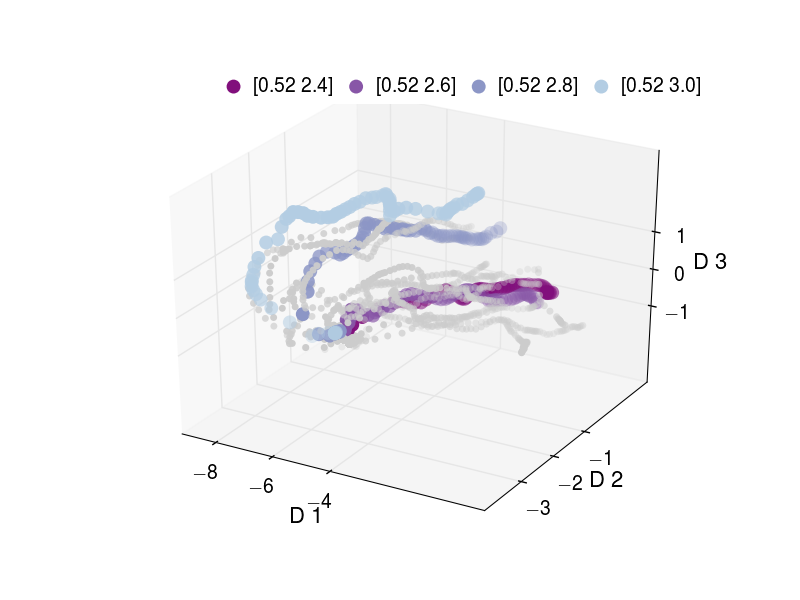}
    \includegraphics[width=.24\textwidth,trim={3cm 1.5cm 1.75cm 1.5cm},clip]{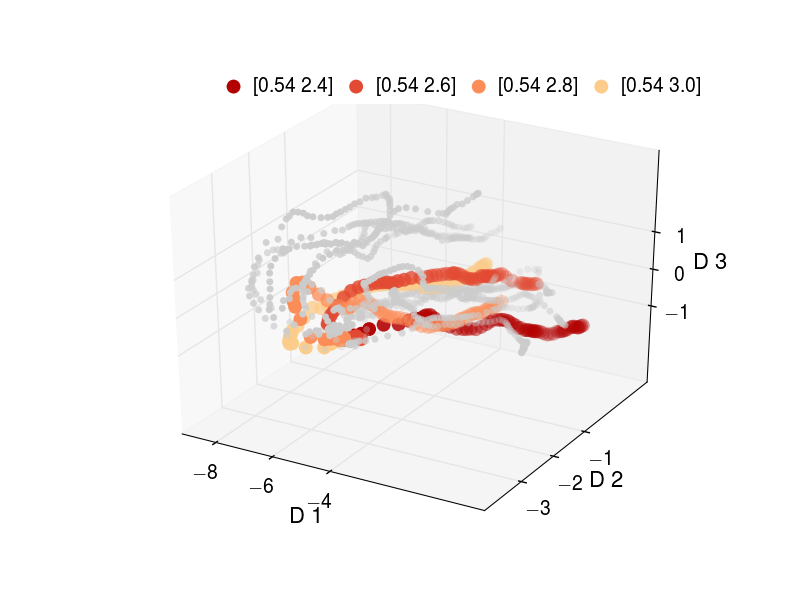}
    \includegraphics[width=.24\textwidth,trim={3cm 1.5cm 1.75cm 1.5cm},clip]{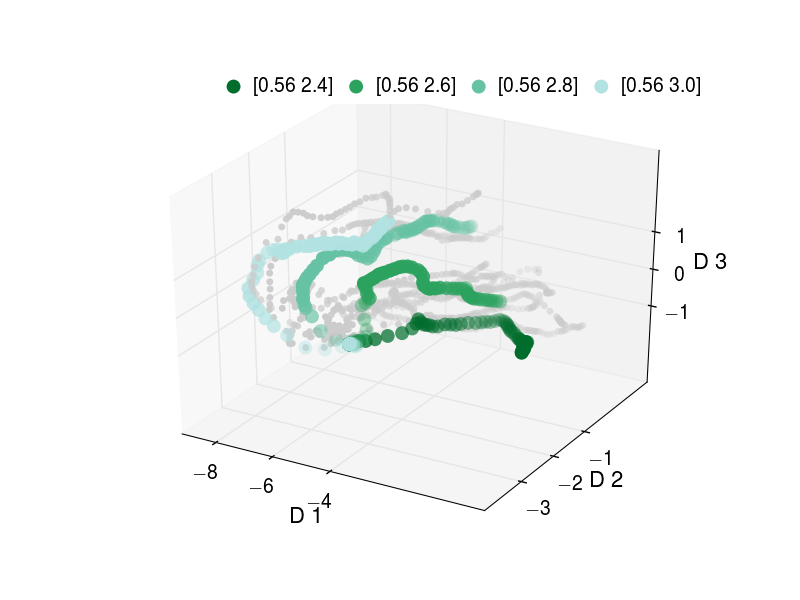}

\includegraphics[width=.24\textwidth,trim={3cm 1.5cm 1.75cm 1.5cm},clip]{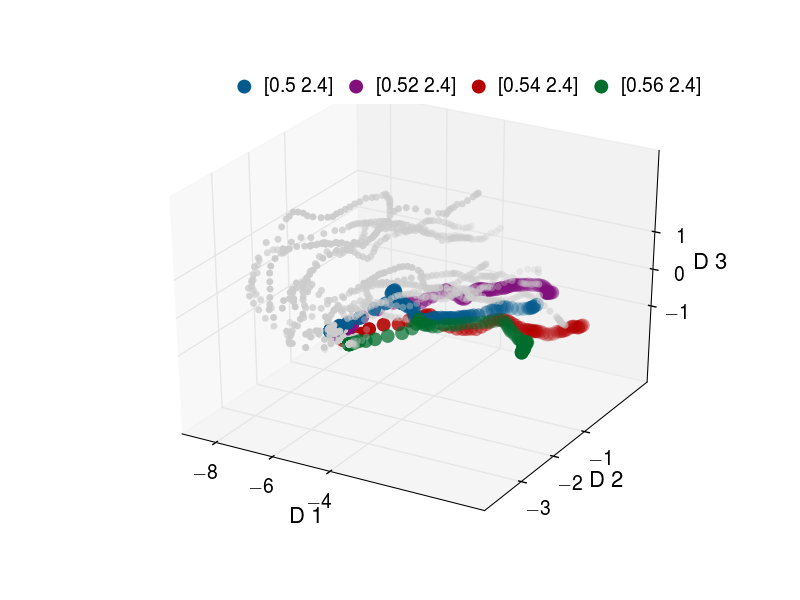}
    \includegraphics[width=.24\textwidth,trim={3cm 1.5cm 1.75cm 1.5cm},clip]{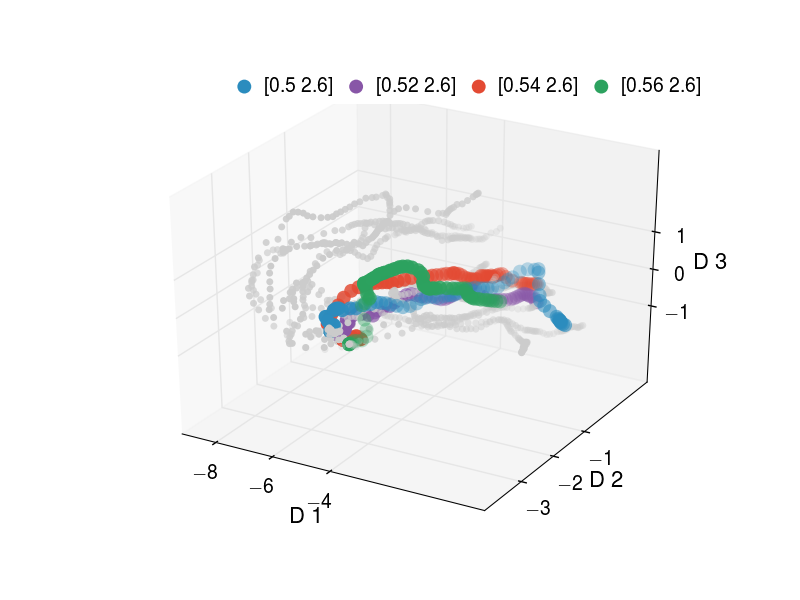}
    \includegraphics[width=.24\textwidth,trim={3cm 1.5cm 1.75cm 1.5cm},clip]{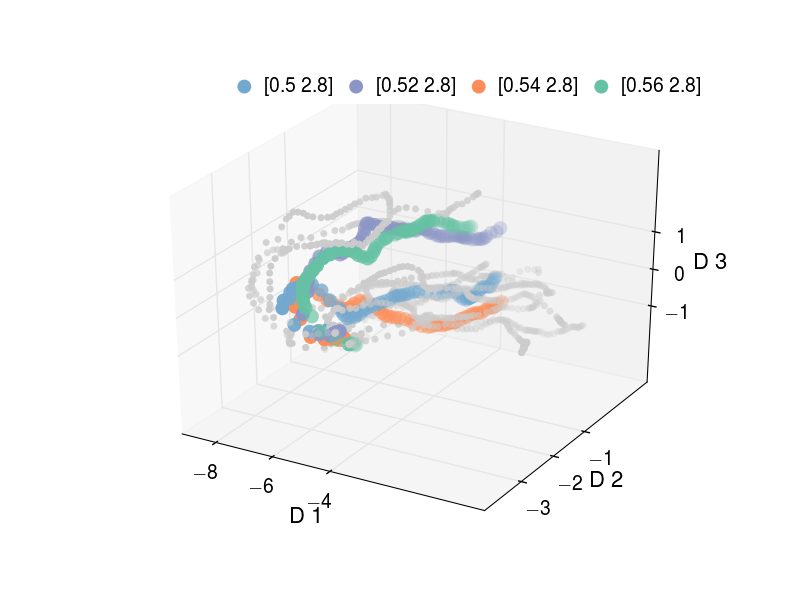}
    \includegraphics[width=.24\textwidth,trim={3cm 1.5cm 1.75cm 1.5cm},clip]{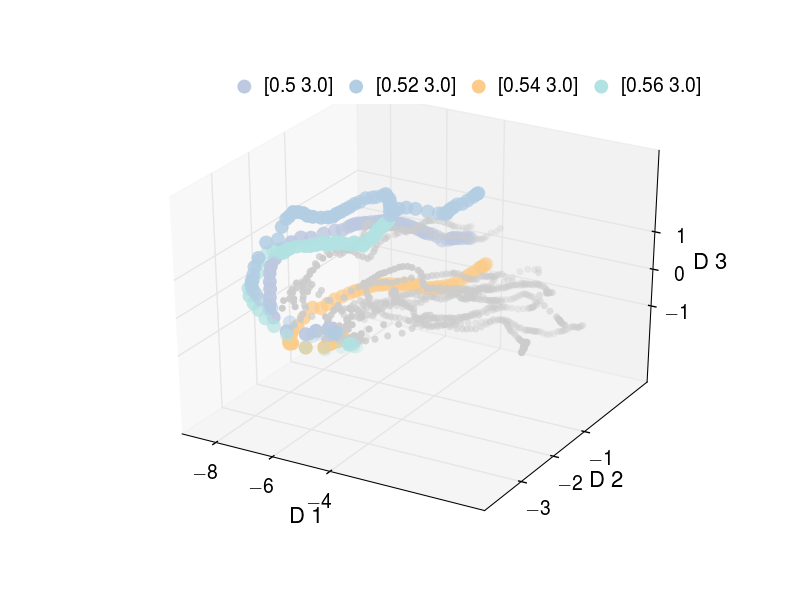}
    \caption{The manifold of the late stage with the first 80 points per each trajectory. The same manifold is color coded by increasing $\phi$ (top) and $\chi$ (bottom). (Please view in color).}    
    \label{results:late}
\end{figure*}

Figures~\ref{results:early} and~\ref{results:late} depict three dimensional manifold discovered using {\sf Entropy-Isomap} for the complete set of 16 pathways. When mapped to the manifold, the pathways show ordering according to the process variables that were varied to generate the data. In both figures, for easier inspection we marked the pathways according to one varying variable. For example, the top row in Fig.~\ref{results:early} depicts the pathways for fixed $\phi$ and varying $\chi$.
Pathways for increasing $\chi$ are ordered from right (dark) to left (light), while pathways for increasing $\phi$ are ordered from front (green) to back (blue).

The observed ordering of pathways strongly indicates that the variables are also latent variables controlling the dynamic process. More importantly, the ordering reveals that denser sampling is required in $\phi$ space. Specifically, the pathways sharing the same $\chi$ but varying $\phi$ are spread further apart than these sharing the same $\chi$ value. This observation has important implications for the design of the next round of exploration in the design space. In particular, the $\phi$ space offers higher exploration benefits, while $\chi$ space has better exploitation chance. This suggests that $\phi$ space should be explored first, followed by potential exploitation phase.

Finally, we notice that {\sf Entropy-Isomap} mapped the data into two regions. The early stages of the process are mapped to evolve in the radial direction, while late stages are mapped parallel to each other. This is interesting as the underlying process indeed has two inherent time scales. In the early stage, the phase separation between two polymer occurs. During this stage, the changes mostly result in increase of the composition amplitude. In the second stage, the coarsening between already formed domains occurs. Here, the amplitude of the composition (signal) does not change significantly. The changes mostly occur in the frequency space with the domain sizes increases over time. 

\section{Conclusions}

Dynamic process data, represented by data trajectories, is challenging to commonly used SDR methods. This is due the strong temporal correlations within data trajectories, that lead to poor quality of recovered manifold. In this work, we introduce the notion of {\em neighborhood entropy}, which quantifies the information exchange between data points in dynamic process data. Then, we present {\sf Entropy-Isomap}, a new algorithm that uses {\em neighborhood entropy} to learn more reliable geodesics, 
and is able to discover latent variables governing dynamic processes, and learn the true manifold~geometry. 

We showcased our method on the data capturing the morphology evolution of materials. The method ordered the trajectory data according to the two process variables. Moreover, it exposed the need for more dense sampling in one of the explored variables. This observation can be used to design the next round of simulations to generate more data for under-sampled process configurations. This demonstrates that the method can be used to guide the data exploration process, potentially reducing the number of required numerical experiments.





\bibliographystyle{IEEEtran}
\bibliography{sample-bibliography.bib}

\begin{thebibliography}{10}
\providecommand{\url}[1]{#1}
\csname url@samestyle\endcsname
\providecommand{\newblock}{\relax}
\providecommand{\bibinfo}[2]{#2}
\providecommand{\BIBentrySTDinterwordspacing}{\spaceskip=0pt\relax}
\providecommand{\BIBentryALTinterwordstretchfactor}{4}
\providecommand{\BIBentryALTinterwordspacing}{\spaceskip=\fontdimen2\font plus
\BIBentryALTinterwordstretchfactor\fontdimen3\font minus
  \fontdimen4\font\relax}
\providecommand{\BIBforeignlanguage}[2]{{%
\expandafter\ifx\csname l@#1\endcsname\relax
\typeout{** WARNING: IEEEtran.bst: No hyphenation pattern has been}%
\typeout{** loaded for the language `#1'. Using the pattern for}%
\typeout{** the default language instead.}%
\else
\language=\csname l@#1\endcsname
\fi
#2}}
\providecommand{\BIBdecl}{\relax}
\BIBdecl

\bibitem{Schoeneman2017}
F.~Schoeneman, S.~Mahapatra, V.~Chandola, N.~Napp, and J.~Zola, ``Error metrics
  for learning reliable manifolds from streaming data,'' in \emph{Proceedings
  of the 2017 SIAM International Conference on Data Mining}.\hskip 1em plus
  0.5em minus 0.4em\relax SIAM, 2017, pp. 750--758.

\bibitem{Tenenbaum2000}
J.~Tenenbaum, V.~de~Silva, and J.~Langford, ``A global geometric framework for
  nonlinear dimensionality reduction,'' \emph{Science}, vol. 290, no. 5500, p.
  2319, 2000.

\bibitem{Lim2003}
I.~Lim, P.~de~Heras~Ciechomski, S.~Sarni, and D.~Thalmann, ``Planar arrangement
  of high-dimensional biomedical data sets by isomap coordinates,'' in
  \emph{Proceedings of the 6th IEEE Symposium on Computer-Based Medical
  Systems}, 2003, pp. 50--55.

\bibitem{Dawson2005}
K.~Dawson, R.~Rodriguez, and W.~Malyj, ``Sample phenotype clusters in
  high-density oligonucleotide microarray data sets are revealed using isomap,
  a nonlinear algorithm,'' \emph{BMC Bioinformatics}, vol.~6, no.~1, p. 195,
  2005.

\bibitem{Zhang2006}
Q.~Zhang, R.~Souvenir, and R.~Pless, ``On manifold structure of cardiac mri
  data: Application to segmentation,'' in \emph{IEEE Computer Society
  Conference on Computer Vision and Pattern Recognition}, vol.~1, 2006, pp.
  1092--1098.

\bibitem{Rohde2008}
G.~Rohde, W.~Wang, T.~Peng, and R.~Murphy, ``Deformation-based nonlinear
  dimension reduction: Applications to nuclear morphometry,'' in \emph{5th IEEE
  International Symposium on Biomedical Imaging: From Nano to Macro}, 2008, pp.
  500--503.

\bibitem{Ruan2014}
Y.~Ruan, G.~House, S.~Ekanayake, U.~Schutte, J.~Bever, H.~Tang, and G.~Fox,
  ``Integration of clustering and multidimensional scaling to determine
  phylogenetic trees as spherical phylograms visualized in 3 dimensions,'' in
  \emph{2014 14th IEEE/ACM International Symposium on Cluster, Cloud and Grid
  Computing (CCGrid)}, 2014, pp. 720--729.

\bibitem{Strange2014}
H.~Strange and R.~Zwiggelaar, \emph{Open Problems in Spectral Dimensionality
  Reduction}.\hskip 1em plus 0.5em minus 0.4em\relax Springer, 2014.

\bibitem{Samudrala2015}
S.~Samudrala, J.~Zola, S.~Aluru, and B.~Ganapathysubramanian, ``Parallel
  framework for dimensionality reduction of large-scale datasets,''
  \emph{Scientific Programming}, 2015.

\bibitem{Jenkins2004}
O.~Jenkins and M.~Matari{\'c}, ``A spatio-temporal extension to isomap
  nonlinear dimension reduction,'' in \emph{Proceedings of the twenty-first
  international conference on Machine learning}.\hskip 1em plus 0.5em minus
  0.4em\relax ACM, 2004, p.~56.

\bibitem{Pearson1901}
K.~Pearson~F.R.S., ``Liii. on lines and planes of closest fit to systems of
  points in space,'' \emph{The London, Edinburgh, and Dublin Philosophical
  Magazine and Journal of Science}, vol.~2, no.~11, pp. 559--572, 1901.

\bibitem{Cox2000}
T.~Cox and M.~Cox, \emph{Multidimensional Scaling, Second Edition}.\hskip 1em
  plus 0.5em minus 0.4em\relax Chapman and Hall/CRC, 2000.

\bibitem{Scholkopf1998}
B.~Sch{\"o}lkopf, A.~Smola, and K.~M{\"u}ller, ``Nonlinear component analysis
  as a kernel eigenvalue problem,'' \emph{Neural computation}, vol.~10, no.~5,
  pp. 1299--1319, 1998.

\bibitem{WodoBaskar2012-CmpMatSc}
O.~Wodo and B.~Ganapathysubramanian, ``Modeling morphology evolution during
  solvent-based fabrication of organic solar cells,'' \emph{Computational
  Materials Science}, vol.~55, pp. 113--126, 2012.

\bibitem{Maaten2008}
L.~van~der Maaten and G.~Hinton, ``Viualizing data using t-sne,'' vol.~9, pp.
  2579--2605, 11 2008.

\bibitem{Roweis2000}
S.~Roweis and L.~Saul, ``Nonlinear dimensionality reduction by locally linear
  embedding,'' \emph{Science}, vol. 290, no. 5500, pp. 2323--2326, 2000.

\bibitem{osc-transistors}
B.~Crone, A.~Dodabalapur, Y.-Y. Lin, R.~Filas, Z.~Bao, A.~LaDuca,
  R.~Sarpeshkar, H.~Katz, and W.~Li, ``Large-scale complementary integrated
  circuits based on organic transistors,'' \emph{Nature}, vol. 403, no. 6769,
  pp. 521--523, 2000.

\bibitem{osc-transistors-review}
C.~D. Dimitrakopoulos and D.~J. Mascaro, ``Organic thin-film transistors: A
  review of recent advances,'' \emph{IBM Journal of Research and Development},
  vol.~45, no.~1, pp. 11--27, 2001.

\bibitem{HS04}
H.~Hoppe and N.~Sariciftci, ``Organic solar cells: An overview,'' \emph{Journal
  of Materials Research}, vol.~19, pp. 1924--1945, 2004.

\bibitem{osc-book}
C.~Brabec, U.~Scherf, and V.~Dyakonov, \emph{Organic photovoltaics: materials,
  device physics, and manufacturing technologies}.\hskip 1em plus 0.5em minus
  0.4em\relax John Wiley \& Sons, 2014.

\bibitem{oled-review1}
Y.-S. Tyan, ``Organic light-emitting-diode lighting overview,'' \emph{Journal
  of Photonics for Energy}, vol.~1, no.~1, pp. 011\,009--011\,009, 2011.

\bibitem{oled-review2}
N.~Thejo~K. and S.~Dhoble, ``Organic light emitting diodes: Energy saving
  lighting technology---a review,'' \emph{Renewable and Sustainable Energy
  Reviews}, vol.~16, no.~5, pp. 2696--2723, 2012.

\bibitem{displays}
G.~Crawford, \emph{Flexible flat panel displays}.\hskip 1em plus 0.5em minus
  0.4em\relax John Wiley \& Sons, 2005.

\bibitem{organic-rfdis}
K.~Myny, S.~Steudel, S.~Smout, P.~Vicca, F.~Furthner, B.~van~der Putten, A.~K.
  Tripathi, G.~H. Gelinck, J.~Genoe, and W.~Dehaene, ``Organic rfid transponder
  chip with data rate compatible with electronic product coding,''
  \emph{Organic Electronics}, vol.~11, no.~7, pp. 1176--1179, 2010.

\bibitem{organic-rfdis2}
K.~Myny, S.~Steudel, P.~Vicca, S.~Smout, M.~J. Beenhakkers, N.~van Aerle,
  F.~Furthner, B.~van~der Putten, A.~K. Tripathi, and G.~H. Gelinck, ``Organic
  rfid tags,'' in \emph{Applications of Organic and Printed Electronics}.\hskip
  1em plus 0.5em minus 0.4em\relax Springer, 2013, pp. 133--155.

\bibitem{smartTextiles}
M.~Stoppa and A.~Chiolerio, ``Wearable electronics and smart textiles: a
  critical review,'' \emph{Sensors}, vol.~14, no.~7, pp. 11\,957--11\,992,
  2014.

\bibitem{organicSkin}
T.~Someya, T.~Sekitani, S.~Iba, Y.~Kato, H.~Kawaguchi, and T.~Sakurai, ``A
  large-area, flexible pressure sensor matrix with organic field-effect
  transistors for artificial skin applications,'' \emph{Proceedings of the
  National Academy of Sciences of the United States of America}, vol. 101,
  no.~27, pp. 9966--9970, 2004.

\bibitem{lochner2014all}
C.~Lochner, Y.~Khan, A.~Pierre, and A.~Arias, ``All-organic optoelectronic
  sensor for pulse oximetry,'' \emph{Nature communications}, vol.~5, 2014.

\bibitem{zhu2014photoreconfigurable}
C.~Zhu, C.~Ninh, and C.~Bettinger, ``Photoreconfigurable polymers for
  biomedical applications: chemistry and macromolecular engineering,''
  \emph{Biomacromolecules}, vol.~15, no.~10, pp. 3474--3494, 2014.

\bibitem{WodoBaskar2011-JCP}
O.~Wodo and B.~Ganapathysubramanian, ``Computationally efficient solution to
  the {C}ahn-{H}illiard equation: adaptive implicit time schemes, mesh
  sensitivity analysis and the 3{D} isoperimetric problem,'' \emph{Journal of
  Computational Physics}, vol. 230, pp. 6037--6060, 2011.

\end{thebibliography}

\end{document}